\documentclass[letterpaper]{article} % DO NOT CHANGE THIS

\usepackage{aaai24}  % DO NOT CHANGE THIS
\usepackage{times}  % DO NOT CHANGE THIS
\usepackage{helvet}  % DO NOT CHANGE THIS
\usepackage{courier}  % DO NOT CHANGE THIS
\usepackage[hyphens]{url}  % DO NOT CHANGE THIS
\usepackage{graphicx} % DO NOT CHANGE THIS
\usepackage{natbib}  % DO NOT CHANGE THIS AND DO NOT ADD ANY OPTIONS TO IT
\usepackage{caption} % DO NOT CHANGE THIS AND DO NOT ADD ANY OPTIONS TO IT
\usepackage{algorithm}
\usepackage{algorithmic}
\usepackage{booktabs}
\usepackage{multirow} 
\usepackage{tabularx}
\usepackage{subcaption}
\usepackage{newfloat}
\usepackage{listings}
\usepackage{amsmath}
\usepackage{amssymb}
\usepackage{xcolor}
\usepackage{bibentry}
\usepackage{afterpage}
\usepackage{natbib}
\setcitestyle{numbers,square}
\usepackage[absolute,overlay]{textpos}

\usepackage{newfloat}
\usepackage{listings}
\DeclareCaptionStyle{ruled}{labelfont=normalfont,labelsep=colon,strut=off} % DO NOT CHANGE THIS
\floatstyle{ruled}
\newfloat{listing}{tb}{lst}{}
\floatname{listing}{Listing}
\title{Learning to Stop Cut Generation for Efficient Mixed-Integer Linear Programming}
\author{
    Haotian Ling, Zhihai Wang, Jie Wang\thanks{Corresponding author.}\\
}
\affiliations{
    University of Science and Technology of China\\
\{haotianling,zhwangx\}@mail.ustc.edu.cn\\
jiewangx@ustc.edu.cn
}

\usepackage{bibentry}
\begin{document}

\maketitle

\begin{abstract}
Cutting planes (cuts) play an important role in solving mixed-integer linear programs (MILPs), as they significantly tighten the dual bounds and improve the solving performance. A key problem for cuts is when to stop cuts generation, which is important for the efficiency of solving MILPs. However, many modern MILP solvers employ hard-coded heuristics to tackle this problem, which tends to neglect underlying patterns among MILPs from certain applications. To address this challenge, we formulate the cuts generation stopping problem as a reinforcement learning problem and propose a novel \textbf{hy}brid \textbf{g}raph \textbf{r}epresentation m\textbf{o}del (\textbf{HYGRO}) to learn effective stopping strategies. An appealing feature of HYGRO is that it can effectively capture both the dynamic and static features of MILPs, enabling dynamic decision-making for the stopping strategies. To the best of our knowledge, HYGRO is \textit{the first} data-driven method to tackle the cuts generation stopping problem. By integrating our approach with modern solvers, experiments demonstrate that HYGRO significantly improves the efficiency of solving MILPs compared to competitive baselines, achieving up to $31\%$ improvement.
\end{abstract}

\section{Introduction}

Mixed-Integer Linear Programming (MILP) is a widely-used mathematical optimization model employed to solve various real-world problems, such as production planning \cite{Pochet2010ProductionPB}, vehicle routing \cite{Laporte2009FiftyYO}, project scheduling \cite{Berthold2010ACI}, and facility location \cite{Abend2016FacilityLC}. A MILP aims to find the values of the decision variables, including integer and continuous, to optimize a linear objective function while satisfying all the linear constraints. A standard MILP takes the form of
\begin{equation}
\label{definition_MILP}
\begin{aligned}
\min_{\mathbf{x}} \{\mathbf{c^T} \mathbf{x}|\mathbf{A}\mathbf{x}\leq \mathbf{b},x_j \in \mathbb{Z},\forall j\in I\} ,
\end{aligned}
\end{equation}
where $\mathbf{c}\in \mathbb{R}^n$,$\mathbf{b}\in \mathbb{R}^m$ and $\mathbf{A}\in\mathbb{R}^{m\times n}$. $\mathbf{x}\in\mathbb{R}^n$ represents a vector of decision variables, where a subset denoted as \(\{x_j|\space\space j\in I, I\subseteq\{1,2,\dots,n\}\}\) consists of integer decision variables, while the remaining decision variables are continuous. MILP solving initiates with its Linear Programming (LP) relaxation, where the integer decision variables in Definition (\ref{definition_MILP}) are treated as continuous. The specific form of LP relaxation can be found in the Definition (\ref{relaxation}).
% are the coefficient vector of the objective function and the constraint coefficient matrix, respectively.
% while the constraints $\mathbf{A}\mathbf{x} +\mathbf{B}\mathbf{y} \leq \mathbf{b}$ define the limits for these variables. The region in which the decision variables $\mathbf{x}$ and $\mathbf{y}$ satisfy all the constraints is called the \textbf{feasible region}. According to the definition (\ref{definition_MILP}), the upper bound on the optimal objective function value obtained during the solving process of the MILP problem is referred to as the \textbf{primal bound}.
 %%割平面流程

A key component in modern MILP solvers is the cutting plane method. The cutting plane method aims to generate valid linear constraints---known as \textbf{cutting planes} (cuts)---that are able to tighten the LP relaxation of the original problem. They are usually repeatedly generated and added to the original problem in multiple rounds, as illustrated in Figure \ref{fig:task}, which has a significant impact on the efficiency of solving MILPs \cite{Achterberg2009, wesselmann2012implementing}. 
%%相关工作 

Recently, using machine learning (ML) to improve the cutting plane method in modern MILP solvers has been an active topic of significant interest. Many existing methods \cite{Tang2020, HUANG2022108353, paulus2022learning, turner2022adaptive, wang2023learning} introduced machine learning into the selection of cuts. Besides, Berthold, Francobaldi, and Hendel \cite{berthold2022learning} proposed to dynamically determine whether to use local cuts. As for the theoretical analysis, Balcan et al.\cite{ theoratical} provided provable guarantees for learning high-performing cuts. They have demonstrated the strong ability to enhance MILP solvers via incorporating ML into the cutting plane method. 

However, the stopping strategy for the cutting plane method---which significantly impacts the efficiency of solving MILPs (see Section \ref{sec:motivation_results})---has received limited attention. On one hand, stopping cuts generation too early often struggles to effectively tighten the LP relaxation, making it difficult for the solver to efficiently find optimal solutions \cite{too_early}. On the other hand, adding too many cuts leads to a significant increase in the size of MILPs, posing a computational problem and thus degrading the solving efficiency \cite{wesselmann2012implementing}. Although many modern solvers employ stopping strategies based on hard-coded heuristics, they often tend to neglect underlying patterns among MILPs from certain applications \cite{BENGIO2021405, gupta2022lookback}. 

To address this challenge, we propose a novel \textbf{hy}brid \textbf{g}raph \textbf{r}epresentation m\textbf{o}del (HYGRO) to learn intelligent stopping strategies based on a novel transformed reinforcement learning formulation. To the best of our knowledge, HYGRO is \textit{the first} data-driven approach to tackle the cuts generation stopping problem, which significantly improves the efficiency of solving MILPs. Specifically, we first transform the stopping strategy learning problem into a reinforcement learning problem to learn simple and efficient stopping conditions, which can significantly reduce the action space and thus improve the performance of learned models (see Section \ref{ablation}). Then, we propose a novel \textbf{hy}brid \textbf{g}raph \textbf{r}epresentation m\textbf{o}del (HYGRO) to learn effective stopping conditions, which captures the underlying patterns among MILPs by integrating their dynamic graph embeddings with their static intrinsic features. We incorporate our proposed HYGRO into the open-source state-of-the-art MILP solver, namely SCIP \cite{BestuzhevaEtal2021OO}. Extensive experiments demonstrate that HYGRO significantly improves the efficiency of solving MILPs, achieving up to $31\%$ improvement, on six challenging MILP problem benchmarks compared to eight competitive baselines. 

We summarize our major contributions as follows. (1) We propose a novel \textbf{hy}brid \textbf{g}raph \textbf{r}epresentation m\textbf{o}del (HYGRO) to learn intelligent stopping strategies based on a novel transformed reinforcement learning formulation. (2) To the best of our knowledge, HYGRO is \textit{the first} approach to learn cuts generation stopping strategies, which significantly improves the efficiency of solving MILPs. (3) Experiments demonstrate that HYGRO significantly improves the efficiency of solving MILPs compared to eight competitive baselines on six challenging MILP benchmarks.

\section{Background}
\subsection{Cutting Plane Method}

The cutting plane method \cite{Gomory} is employed by modern MILP solvers to iteratively add valid cutting planes to the original MILP problem. Let's start by considering the LP relaxation of the MILP problem. where we treat the decision variables in Definition (\ref{definition_MILP}) as continuous, leading to:
\begin{equation}
\label{relaxation}
    \min_{\mathbf{x}} \{\mathbf{c}^{\mathbf T} \mathbf{x}|\mathbf{A}\mathbf{x} \leq \mathbf{b},\mathbf{x}\in \mathbb{R}^n\},
\end{equation}
where $\mathbf{x}$ represents a vector of continuous decision variables. The optimal value of the LP relaxation is termed the \textbf{dual bound}. Specifically, if we find an optimal solution $\mathbf{x}^{lp}$ for the relaxation problem, the term $\mathbf{c}^{\mathbf{T}}\mathbf{x}^{lp}$ is referred to as the dual bound. Besides, the optimal achievable objective function value during the solving process of the original MILP problem is called the \textbf{primal bound}. More precisely, denoting the current optimal solution that satisfies all constraints as $\mathbf{x}^{*}$, then $\mathbf{c}^{\mathbf{T}}\mathbf{x}^{*}$ represents the primal bound.

In the cutting plane method, valid inequalities are generated from the constraints of the original MILP problem using algorithms like Gomory Cuts \cite{Gomory}, added to strengthen the relaxation \cite{Achterberg2009}, resulting in tighter LP relaxation. The process of cut generation and addition \textbf{(G\&A)} is performed iteratively until a stopping condition is triggered. A cut is represented as: $\sum_{j \in S} a_j x_j \leq d$, where $a_j$ denotes the coefficients in the cut, $S$ represents a set of decision variable indices, and $d$ is the right-hand side of the cut. The set of all cuts generated in a single iteration is denoted as $\mathcal{C}$. From $\mathcal{C}$, we select a subset $\mathcal{C^*}$ to be added to the original MILP problem. Commonly, in modern solvers like SCIP \cite{BestuzhevaEtal2021OO}, a method called ``Branch and Cut" (\textbf{B\&C}) \cite{branch_cut} is employed.

\begin{figure}[t]
    \centering
    \includegraphics[width=0.98\linewidth]{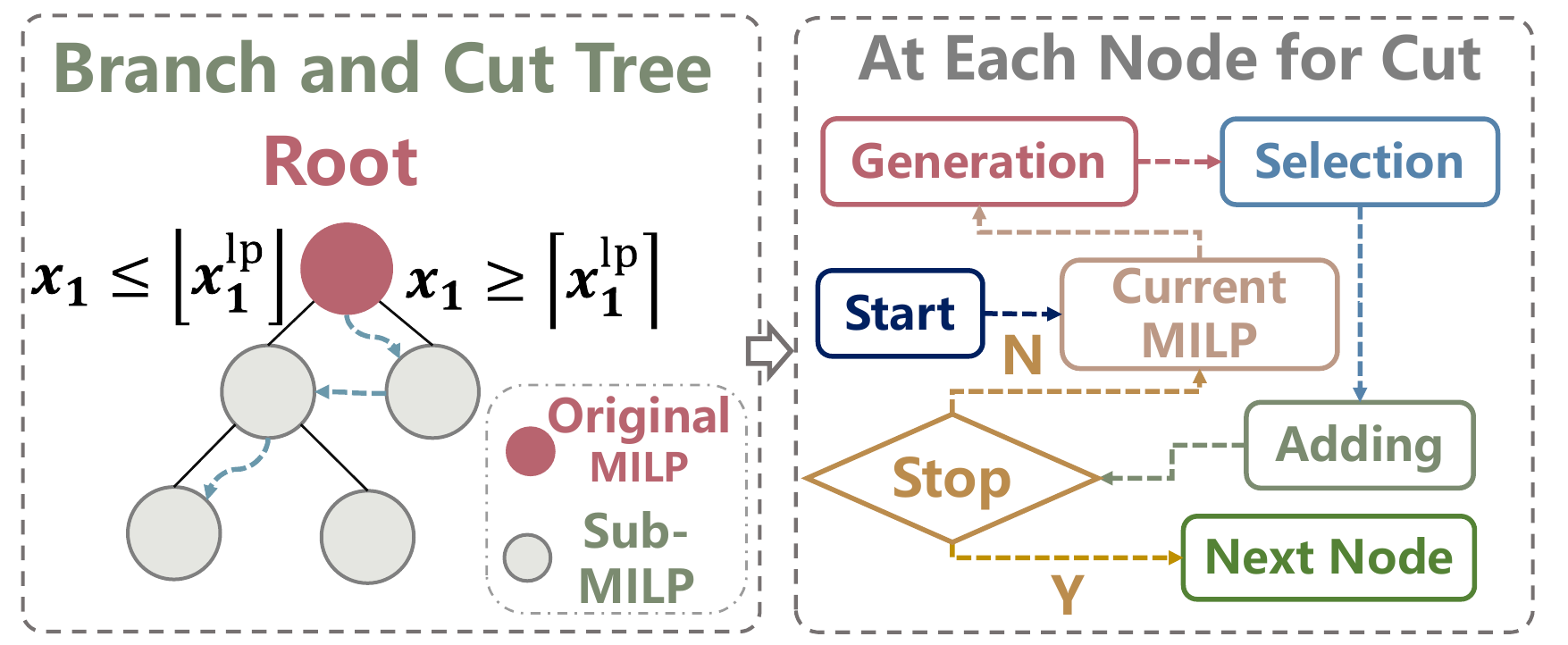}
    \caption{The right gray dashed box illustrates main steps related to cutting planes that each node needs to undergo.}
    \label{fig:task}
\end{figure}
\subsection{Stopping Strategies in Cutting Plane Method }\label{Heur_Stopping_Strategy}

We denote $\pi_{\mathcal{S}}$ as a cutting plane stopping strategy. In modern solvers, within a B\&C tree of an MILP instance, at node $i$, the G\&A of cutting planes occurs iteratively. In each iteration, $\pi_{\mathcal{S}}$ decides whether to proceed to the next iteration. If an iteration is interrupted by $\pi_{\mathcal{S}}$, the process will move to the next node $j$. Related illustrations can be found in Figure \ref{fig:task}. Here are several classical heuristic stopping strategies.

\subsubsection{Fixed Cut Numbers (FCN)} \label{FixedNumber}
The strategy will stop the cut generation after a specific number of cuts have been added, ensuring that the total number of added cuts doesn't exceed the predefined limit. Assuming we set the number of cuts to be added as $k$, then at each node, we add a maximum of $k$ cuts. Specifically, let $C_i$ represent the set of cuts added at node $i$, then the FCN will have \(|C_i| \leq k ,\) where $|C_i|$ represents the number of cuts added at node $i$.

\subsubsection{Fixed Cut Rounds (FCR)}\label{FixedRound} 
SCIP uses a round-based approach to add multiple cuts to the original problem. Under the fixed cutting plane rounds strategy, a finite number of rounds for cut generation are executed at each node. Assuming we set a threshold $t$, then at each node, we will perform a maximum of $t$ rounds for cut generation. To be precise, denoting the actual round for cut generation at node $i$ as $t_i$, this stopping strategy follows the constraint: \(t_i \leq t\).

\subsubsection{Stagnation Round Detection (SRD)}\label{StagnationRound}
The two aforementioned strategies have the advantage of simplicity, yet their drawbacks become apparent due to their inability to consider the dynamic factors during the solving process.

Let $C_i^t$ be the set of cutting planes generated in the $t^{th}$ round at node $i$. Define $\mathcal{O^*}(C_i^t)$ as the objective value resulting from the $t^{th}$ round of cutting plane generation. Furthermore, a threshold $\epsilon$ is set to determine whether the objective value improvement is significant. If we have:
\begin{equation}
\label{stagnation_inequality}
\frac{\lvert\mathcal{O^*}(C_i^t) - \mathcal{O^*}(C_i^{t+1})\rvert}{\lvert\mathcal{O^*}(C_i^t)\rvert} \leq \epsilon ,
\end{equation}
then we consider that the solving process has stagnated at the $(t+1)^{th}$ round. We define the number of consecutive stagnation rounds at node $i$ in the round $t^{th}$ round as $s_i^t$, and we set $s_i^0=0$. If the stagnation persists into the $(t+2)^{th}$ round, we update $s_{i}^{t+2}=s_{i}^{t+1}+1$; otherwise, $s_{i}^{t+2}=0$. Let $s_i$ be the maximum allowed consecutive stagnation rounds at node $i$. The stagnation round detection strategies is formulated as: $s_i^r \leq s_i$, where $r$ represents any given round.
\section{Motivation Results}\label{sec:motivation_results}

Empirically, different stopping strategies significantly affect the solving performance of MILP problems. We conducted carefully designed experiments to evaluate this impact.

\subsection{Setup}
To visually demonstrate the impact of stopping strategies on the solving efficiency of MILPs using the cutting plane method, we conducted exhaustive search experiments within a specified range on three datasets: Anonymous, MIK, and Knapsack, using the SCIP solver. For detailed descriptions of the datasets, please refer to Section \ref{Datasets}. We employ the Fixed Cut Round strategy, adding cutting planes only at the root node for each instance, with a fixed number of iterations as mentioned in Section \ref{FixedRound}. For each instance, we perform 100 runs. In the \(j^{th}\) run, for every instance \(\mathcal{I}\), we execute a maximum of \(j\) iterations for the G\&A of cutting planes at the root node. The Anonymous dataset employs primal-dual bound gap integral (PDI) \cite{wang2023learning}, while other datasets represent solving performance using solving time.

\subsection{Results}
\begin{figure}[t]
\centering
    \includegraphics[width=0.49\linewidth]{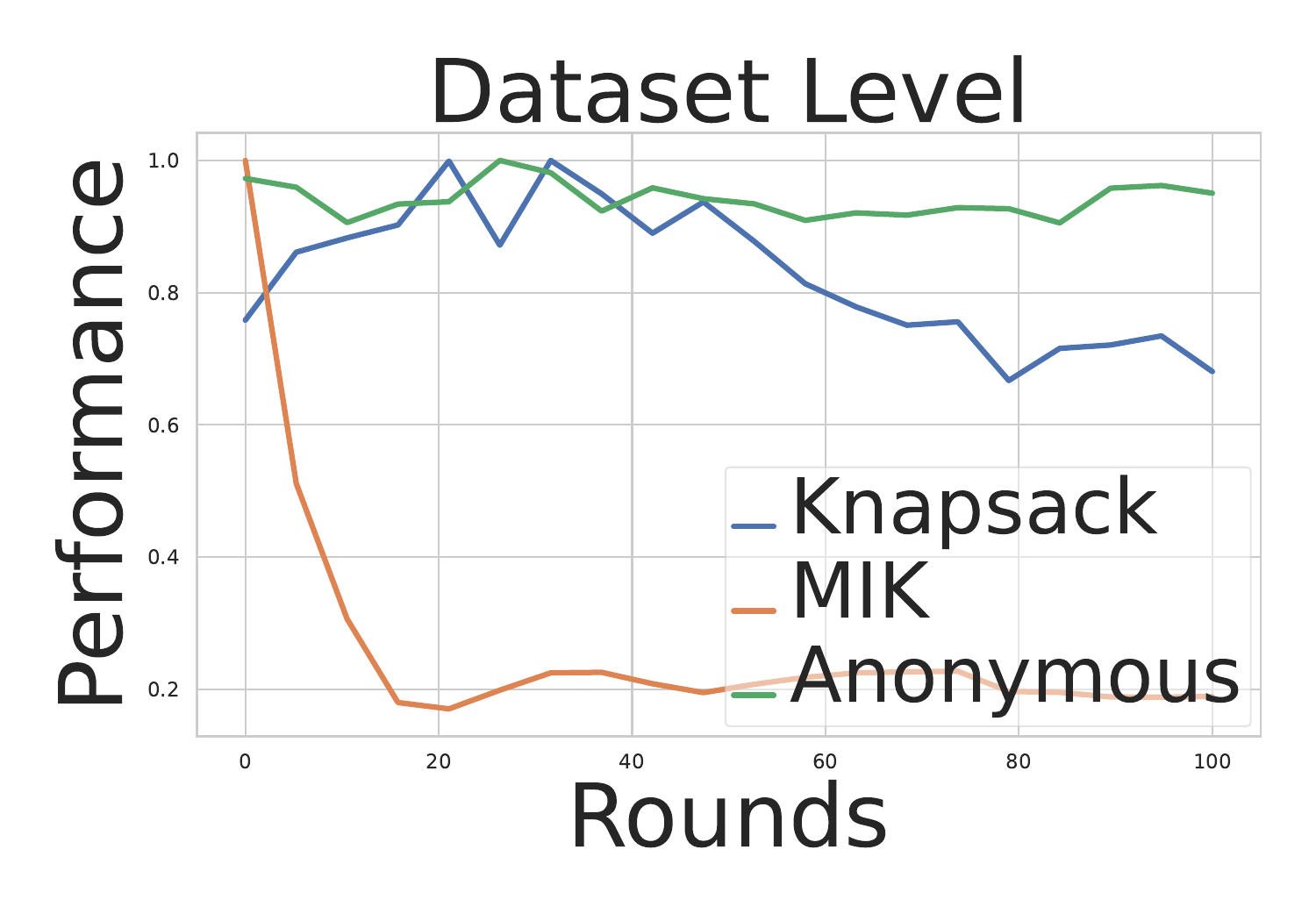}
    \includegraphics[width=0.49\linewidth]{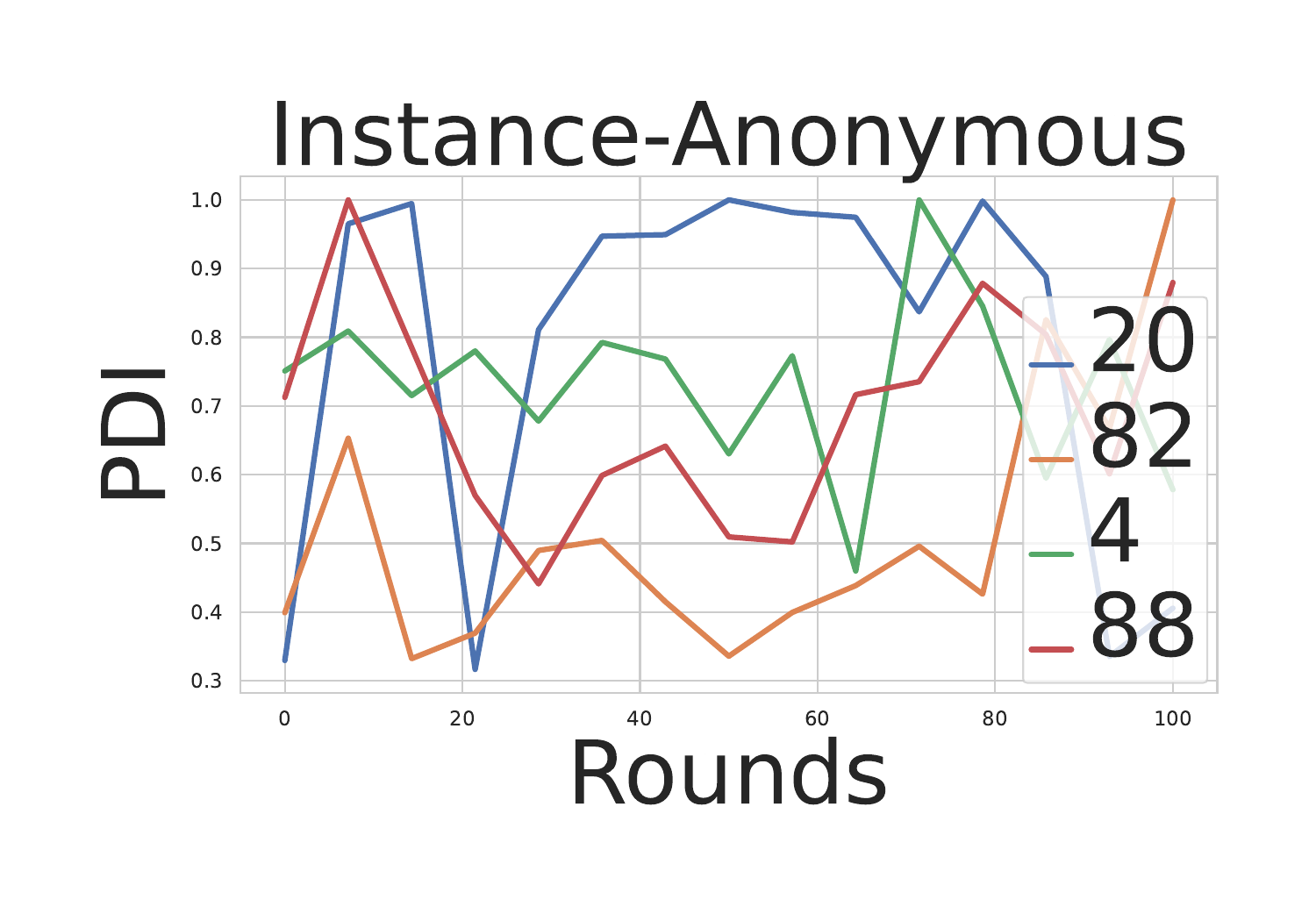}
\caption{Lower values indicate better performance on the figures. Data for both figures was normalized. The legend in the right figure represents the instance numbers in the Anonymous dataset.}
\label{fig:motivation}
\end{figure}

From Figure \ref{fig:motivation} (Left), it is evident that as the number of cutting plane G\&A rounds varies, the trend in solving performance varies across different datasets. This indicates that different types of MILP problems need diverse stopping strategies to achieve enhanced performance. Furthermore, Figure \ref{fig:motivation} (Right) demonstrates that even within a single dataset, the performance trends across different instances vary significantly, which emphasizes the requirement for instance-specific stopping strategies instead of solely relying on predefined hardcoded strategies. The analyses conducted on both the dataset level and individual MILP instances clearly demonstrate the importance of using learning-based methods to develop more effective stopping strategies.

\section{Method}
\subsection{Motivation\label{transformation}}
Our primary task involves dynamically determining the optimal moment to stop the G\&A of cutting planes at each node within the B\&C tree, enabling efficient solving of MILP. Achieving this goal necessitates a more intelligent stopping strategy. A more intuitive approach is to allow the model to decide, after each G\&A of cutting planes, whether to proceed to the next iteration. However, this method presents two notable drawbacks. \textbf{Firstly}, direct decision-making regarding stopping would result in an increased frequency of invoking the trained model, resulting in elevated computational costs that cannot be disregarded \cite{Hybrid}. \textbf{Additionally}, if the model were to directly determine whether to stop, given a maximum of \(k\) iterations of cutting plane G\&A at the current node, the potential action space would exponentially increase to \(2^k\), making it challenging for the model to learn effective stopping strategies.

To address the aforementioned challenges, we innovatively transformed our approach, shifting from directly learning whether to stop to indirectly learning simple and efficient stopping conditions. This transformation draws inspiration from the SRD strategy mentioned in Section \ref{StagnationRound}, incorporating prior knowledge that if cutting planes are unable to effectively tighten the LP relaxations, their G\&A should be stopped. Specifically, at each node, the model learns to dynamically determine the stopping condition, which is a threshold for the maximum number of consecutive stagnation rounds, instead of relying on a predetermined value. The above transformation significantly improves the performance of learned models while also greatly lowering computational overhead. The effectiveness of this transformation is demonstrated through experiments in Section \ref{ablation}.

\subsection{Problem Formulation}\label{Formulation}
%%v 8.14
Reinforcement learning excels at decision-making tasks, achieving a series of successes \cite{Zhou_AAAI,yang2022learning,SAC,Wang_Zhou_AAAI_2022,RAEB,kuang,Liu_AAAI2023,NEURIPS2020_Zhou} and finding increasing applications in specific tasks \cite{Adriaensen2022AutomatedDA}. We consider the stopping strategy as a reinforcement learning problem as well.

To learn the aforementioned stopping strategy, we formulate the stopping problem as a Markov Decision Process (MDP) \cite{Bellman1957AMD,Sutton}. Specifically, we define the state space \(\mathcal{S}\), action space \(\mathcal{A}\), reward function \(r\), state transition function \(f\), and terminal state \(\mathcal{T}\) as follows.

\subsubsection{State Space}\label{StateSpace}

\begin{figure}[t]
  \centering
  \includegraphics[width=0.95\linewidth]{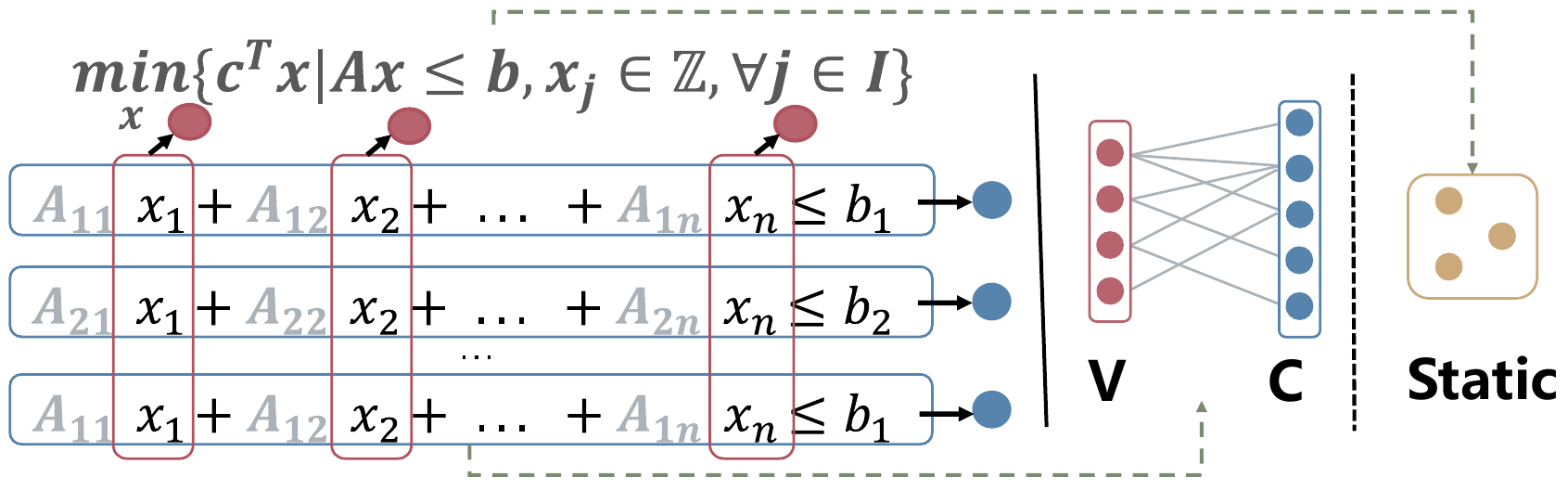}
  \caption{The red, blue, and yellow circles represent variable, constraint, and static features, respectively. }
  \label{fig:features}
\end{figure}

We employ a four-tuple \(\{ \mathcal{C}, \mathcal{V}, \mathcal{E}, \mathcal{S^*} \}\) to denote the state space \(\mathcal{S}\) of the current node, where \(\mathcal{C}\), \(\mathcal{V}\), and \(\mathcal{E}\) correspond to the dynamic part of \(\mathcal{S}\), capturing the features of constraints, variables, and coefficients, respectively. We model these dynamic components using a bipartite graph. \(\mathcal{S^*}\) signifies the static part, consisting of the intrinsic attributes of the MILP problem, and we represent these attributes using a vector. The schematic diagram illustrating these four components is presented in Figure \ref{fig:features}, and their specific forms will be elaborated upon in Section \ref{HYGRO}.

\begin{figure*}[t]
  \centering
  \includegraphics[width=\linewidth, clip]{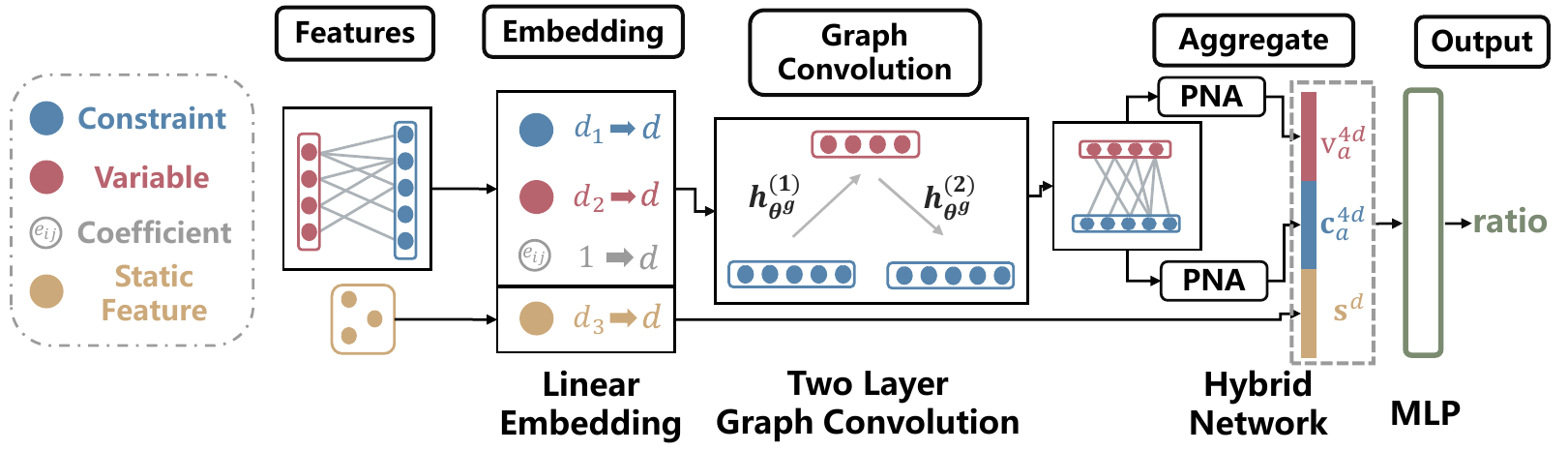}
  \caption{Diagram illustrating the structure of HYGRO. First, it employs a two-layer graph neural convolutional network to encode the dynamic part. Then, it aggregates the dynamic part with the static part and finally outputs the \(ratio\). In summary, HYGRO takes the four features shown in the diagram as inputs and generates the \(ratio\) value for the current node as output.}
  \label{fig:framework}
\end{figure*}
\subsubsection{Action Space}\label{action}
The formulation of our action space, represented as $\mathcal{A}$, is based on the discussions regarding method transformation in Section \ref{transformation}. At node $i$, a single action $a_i \in [0,t_{\mathcal{A}})$ is taken, where $a_i$ is an integer representing the learned stopping condition, i.e., maximum consecutive stagnation rounds for node $i$. The value of $t_{\mathcal{A}}$ corresponds to the size of the action space $\mathcal{A}$. This action $a_i$ guides the solver to iteratively generate and add cutting planes at node $i$ until the stopping condition indicated by $a_i$  is triggered.

\subsubsection{State Transition}
A state transition takes place when the number of consecutive stagnation rounds at node \(i\) exceeds the threshold set by action $a_i$, or when other specific conditions are triggered at node \(i\). The state transition function \(f_t\) outlines the process of transiting from state \(\mathcal{S}_i\) at node \(i\) to state \(\mathcal{S}_j\) at node \(j\) within the Branch and Cut tree: \(\mathcal{S}_j = f_t(\mathcal{S}_i, a_i)\). As illustrated in Figure \ref{fig:task}, the blue dashed curved line indicates the successor in the state transition, which corresponds to the node where the subsequent iteration of cutting plane G\&A takes place.

\subsubsection{Terminal State}
Upon completing the full solving process for the instance $\mathcal{I}$, which could involve achieving the optimal feasible solution or reaching a predefined time limit, we define this occurrence as a transition to the terminal state $\mathcal{T}$.

\subsubsection{Reward}
For a given MILP instance $\mathcal{I}$, the agent performs sequential decision-making at multiple nodes, generating a sequence of actions. Before transiting to the terminal state, the immediate rewards for each action are set to 0. Upon transiting to the terminal state, we compute the reward signal $r$ to holistically assess the quality of these actions. Here, a value $v$ quantifies the effectiveness of these actions, associated with solving performance metrics, such as solving time and PDI. The exact approach for computing the reward signal will be elaborated upon in Section \ref{Trainning}.

%%[0 10.8 0 2.5]

\subsection{Learning to Stop}

In this section, we will present the framework of HYGRO and our strategy for training HYGRO.

\subsubsection{Hybrid Graph Representation Model\label{HYGRO}} 
%%交代特征

An MILP problem \(\mathcal{I}\), corresponding to a node \(i\) in the B\&C tree, has underlying patterns that can be represented using the four-tuple \{\({\mathcal{C}, \mathcal{V}, \mathcal{E}, \mathcal{S^*}}\)\}. The bipartite graph comprising \(\mathcal{C}\), \(\mathcal{V}\), and \(\mathcal{E}\) is an effective model for representing MILP problems, widely employed in MILP research \cite{Gasse2019ExactCO}. In fact, static intrinsic attributes $\mathcal{S^*}$, such as the scale of \(\mathcal{I}\), also play a vital role in the node-level representation of MILP \cite{berthold2022learning}. In order to comprehensively encode $\mathcal{I}$, we propose a novel HYGRO that integrates dynamic graph embeddings with static intrinsic features to learn efficient stopping conditions.

Specifically, assuming there are \(n\) constraints each with \(d_1\) features, \(m\) variables each with \(d_2\) features, and \(e\) non-zero coefficients, we can represent the \{\({\mathcal{C}, \mathcal{V}, \mathcal{E}}\)\}. using matrices \(\mathbf{C}^{n\times d_1}\), \(\mathbf{V}^{m\times d_2}\), and \(\mathbf{E}^{e\times 1}\), collectively forming the bipartite graph of Figure \ref{fig:features}. The bipartite graph formed by these three components undergoes internal interactions, thus we term it the \textbf{dynamic part}. As for intrinsic attributes of \(\mathcal{I}\), such as its scale, total \(d_3\) in count. To represent these attributes, we employ a vector \(\mathbf{s}^{d_3}\). Since \(\mathbf{s}^{d_3}\) remains independent throughout the state encoding procedure, we term it the \textbf{static part}. Then we proceed to perform a linear transformation on \{\({\mathbf{C}^{n\times d_1}, \mathbf{V}^{m\times d_2}, \mathbf{E}^{e\times 1}, \mathbf{s}^{d_3}}\)\}. This transformation uniformly embeds these four components into a $d$-dimensional space, resulting in \{\({\mathbf{C}^{n\times d}, \mathbf{V}^{m\times d}, \mathbf{E}^{e\times d}, \mathbf{s}^{d}}\)\}. 

\newcommand{\avg}{\mathrm{avg}}
\newcommand{\std}{\mathrm{std}}

For the bipartite graph part, we define the graph convolution \cite{GraphConvolutional,NIPS2016_graph} operation as : \(h_{\theta^{g}}^{(l)}(\mathbf{left},\mathbf{right},\mathbf{edge})\), where \(\theta^{g}\) represents the parameters of $h$, \(h\) denote the information propagation between the two sides on layer \(l\), and \(\mathbf{edge}\) represents the associations between \(\mathbf{left}\) and \(\mathbf{right}\). Then, the flow of information can be described as: \(h_{\theta^{g}}^{(1)}(\mathbf{C}^{n\times d},\mathbf{V}^{m\times d},\mathbf{E}^{e\times d})\), \(h_{\theta^{g}}^{(2)}(\mathbf{V}^{m\times d},\mathbf{C}^{n\times d},\mathbf{E}^{e\times d})\). Then we leverage the Principal Neighbourhood Aggregation (PNA) \cite{corso2020principal} to aggregate the \(\mathbf{C}^{n\times d}\) and \(\mathbf{V}^{m\times d}\) into vectors \(\mathbf{c}^{4d}_a\) and \(\mathbf{v}^{4d}_a\) respectively. Specifically, let \([\cdot]\) represent the concatenation between different vectors, our aggregation method can be described as:
\[
\mathbf{d}^{4d}_a=[\max(\mathbf{D}^{q\times d}),\min(\mathbf{D}^{q\times d}),\avg(\mathbf{D}^{q\times d}),\std(\mathbf{D}^{q\times d})],
\]
where the four terms in \([\cdot]\) respectively denote taking the maximum value, the minimum value, the average value, and the standard deviation along the columns of  matrix \(\mathbf{D}^{q\times d}\). 

Finally, we employ a Multilayer Perceptron (MLP) \cite{Rosenblatt1963PRINCIPLESON}, denoted as \(\mathcal{M}\), serving as the output module, we use:
\(ratio = 0.5*\sigma (\mathcal{M}([\mathbf{v}^{4d}_a,\mathbf{c}^{4d}_a,\mathbf{s}^{d}]))+0.5\) to compute the output scalar \(ratio\),
where \(\sigma\) is the \(tanh\) activation function \cite{tanh, tanh2}, mapping any real number to the interval (-1, 1). Thus, we will have \(ratio \in (0, 1)\). The HYGRO's architecture is illustrated in Figure \ref{fig:framework}. Additionally, if the current node is not the root node, then \(ratio = ratio \times \gamma\), where \(\gamma\) is a constant less than 1. This is based on prior knowledge that cutting planes added at the root node are often more critical. Finally, we let the action at node $i$ to be $a_i = \lfloor ratio\cdot t_{\mathcal{A}} \rfloor$, where $\lfloor \cdot \rfloor$ represents the floor operation, $t_\mathcal{A}$ represents the size of the action space.

\subsubsection{Training\label{Trainning}}
We use the Evolutionary Strategies (ES) \cite{Rechenberg1973EvolutionsstrategieO} to train HYGRO. We opted for ES due to its simplicity in training, circumventing the need for explicit gradient computations, enabling us to exploit training parallelism to its fullest extent \cite{2017evolution}. We treat the training process of HYGRO as a black-box optimization problem, where we input the parameters $\theta_0$ of HYGRO and obtain perturbations $\mathbf{\varepsilon}$ along with the corresponding reward signals $\mathbf{r}$ induced by them \cite{Tang2020}. Specifically, for the HYGRO $\mathcal{H}_0$, we generate $k$ perturbations $\mathbf{\varepsilon}$: $\{\epsilon_1, \epsilon_2, ..., \epsilon_k\}$, and then introduce them into $\mathcal{H}_0$ to create: $\{\mathcal{H}^{\epsilon_1}_0, \mathcal{H}^{\epsilon_2}_0, ..., \mathcal{H}^{\epsilon_k}_0\}$. Subsequently, we integrate these new models into the SCIP solver respectively, obtaining the solving performance, such as solving time, \(\mathbf{p}\) of each new model: $\{p_1, p_2, ..., p_k\}$. Then, we compute the reward $r_i$ for each perturbation $\epsilon_i$ using $\mathbf{r} = \text{softmax}(\mathbf{p})$, leveraging \(\mathbf{r}\) and  $\mathbf{\varepsilon}$ to approximate implicit gradient \(\hat{g_\theta}\), iteratively adjusting the parameters $\theta_0$ of HYGRO.

\section{Experiment}

We designed experiments to evaluate the effectiveness of our approach and assess the efficiency of stopping conditions learned by HYGRO. Our experiments mainly focus on the following four aspects. Here is a brief introduction:
\textbf {(1) Solving Performance:} We conducted experiments comparing the solving performance of our method with eight competitive baselines on six NP-hard datasets.
\textbf{(2) Ablation Study:}
As mentioned earlier in Section \ref{transformation}, we designed ablation experiments to validate the effectiveness of the transformation.
\textbf{(3) Generalization Study:} We evaluate the ability of HYGRO to generalize across different sizes of MILPs.
\textbf{(4) Visualization:}
We compared the solving performance achieved by the HYGRO with the optimal performance obtained through an exhaustive search and depicted the results.
% The Set Covering problem dataset consists of 500 constraints and 1000 variables. The MIS dataset involves 500 variables and approximately 2000 constraints. Lastly, the Knapsack Problem dataset consists of 1000 variables and approximately 100 constraints.

\begin{table*}[]

\resizebox{\textwidth}{!}{

\fontsize{60}{66}\selectfont
\begin{tabular}{lcccccccccccc}
\toprule[6pt]
              & \multicolumn{11}{c}{\textbf{Classical Dataset}}                                                                                                                                                                                                                                                                                                   \\ \hline
              & \multicolumn{3}{c}{Set Covering (1000V, C=500)}                                          &  & \multicolumn{3}{c}{Max Independent Set (500V, C\scalebox{2}{$\approx$}2000)}                              &  & \multicolumn{3}{c}{Knapsack (1000V, C\scalebox{2}{$\approx$}200)}                                                  & \multicolumn{1}{c}{\multirow{2}{*}{\begin{tabular}[c]{@{}c@{}}Imprv.\\ Avg.(\%)\end{tabular}}} \\ \cline{1-4} \cline{6-8} \cline{10-12}
Method        & Time(s)\scalebox{2}{$\downarrow$}   & PDI\scalebox{2}{$\downarrow$}          & Imprv.(\%)\scalebox{2}{$\uparrow$}  &  & Time(s)\scalebox{2}{$\downarrow$}  & PDI\scalebox{2}{$\downarrow$}       & Imprv.(\%)\scalebox{2}{$\uparrow$}  &  & Time(s)\scalebox{2}{$\downarrow$}      & PDI\scalebox{2}{$\downarrow$}           & Imprv.(\%)\scalebox{2}{$\uparrow$}  & \multicolumn{1}{c}{}                                                                            \\ \hline
HYGRO*   & \textbf{ 3.91(3.0) } & \textbf{41.94(27.2)}     & \textbf{24.08}        &  & \textbf{5.78(3.8)}  & \textbf{31.75(21.5)} & \textbf{10.53 }       &  & \textbf{9.12(12.6)}     & \textbf{17.67(14.3)}      & \textbf{31.63 }       & \textbf{22.08 }                                                                                               \\
HYGRO I* & \textbf{3.99(2.3)}   & 47.75(26.4)              & \textbf{22.52 }       &  & \textbf{5.61(3.4)}  & \textbf{29.18(19.9)} & \textbf{ 13.16 }      &  & 10.34(12.5)             & 19.23(16.3)               & 22.48                 &\textbf{19.39}                                                                                                 \\ \hline
Default       & 5.15(3.5)            & 55.23(29.4)              & N/A                   &  & 6.46(4.1)           & 33.62(23.4)          & N/A                   &  & 13.34(15.6)             & 22.17(18.9)               & N/A                   & N/A                                                                                                \\
No Cuts       & 4.04(2.8)            & \textbf{43.86(26.1)}     & 21.55                 &  & 9.71(8.4)           & 74.29(59.8)          & -50.31                &  & 11.56(13.5)             & 20.14(16.8)               & 13.34                 &-5.14                                                                                                 \\
Always   & 9.56(4.9)            & 89.01(50.7)              & -85.63                &  & 6.64(4.5)           & 34.44(23.5)          & -2.79                 &  & 40.25(49.3)             & 37.27(22.6)               & -201.72               &-96.71                                                                                                 \\
FCN        & 9.58(4.9)            & 87.84(51.5)              & -86.02                &  & 6.53(4.5)           & 32.33(21.8)          & -1.08                 &  & 32.74(37.8)             & 34.39(18.7)               & -145.43               &-77.15                                                                                                \\
FCR        & 9.81(5.2)            & 88.31(49.4)              & -90.48                &  & 6.66(4.3)           & 33.14(22.7)          & -3.10                 &  & 16.25(16.0)             & 23.84(14.2)               & -21.81                &-38.46                                                                                                 \\
Immediate     & 7.10(4.1)            & 70.64(40.3)              & -37.86                &  & 6.63(4.3)           & 35.46(25.2)          & -2.63                 &  & \textbf{7.74(13.3) }    & \textbf{16.89(10.8)}      & \textbf{ 41.97 }      &0.49                                                                                                 \\
Random I      & 9.08(5.0)            & 84.07(48.6)              & -76.31                &  & 7.04(6.8)           & 37.52(39.6)          & -8.98                 &  & 28.25(28.9)             & 26.57(13.8)               & -111.77               &-65.69                                                                                                 \\
Random II     & 8.56(4.8)            & 81.85(49.5)              & -66.21                &  & 6.81(4.4)           & 35.73(24.2)          & -5.42                 &  & 13.93(15.4)             & 19.05(12.1)               & -4.42                 &-25.35                                                                                                 \\ \toprule[6pt]
              & \multicolumn{11}{c}{\textbf{Harder Dataset}}                                                                                                                                                                                                    &                                                                                                 \\ \hline
              & \multicolumn{3}{c}{Corlat}                                                &  & \multicolumn{3}{c}{MIK}                                              &  & \multicolumn{3}{c}{Anonymous}                                                 & \multicolumn{1}{c}{\multirow{2}{*}{\begin{tabular}[c]{@{}c@{}}Imprv.\\ Avg.(\%)\end{tabular}}} \\ \cline{1-4} \cline{6-8} \cline{10-12}
Method        & Time(s)\scalebox{2}{$\downarrow$}   & PDI\scalebox{2}{$\downarrow$}          & Imprv.(\%)\scalebox{2}{$\uparrow$}  &  & Time(s)\scalebox{2}{$\downarrow$}  & PDI\scalebox{2}{$\downarrow$}       & Imprv.(\%)\scalebox{2}{$\uparrow$}  &  & Time(s)\scalebox{2}{$\downarrow$}      & PDI\scalebox{2}{$\downarrow$}           & Imprv.(\%)\scalebox{2}{$\uparrow$}  & \multicolumn{1}{c}{}                                                                            \\ \hline
HYGRO*   & \textbf{37.7(77.5)} & 1235(2893)          & \textbf{17.3}        &  & \textbf{9.5(14.0)} & \textbf{59.0(82.8)} & \textbf{7.4}         &  & 241(102)           & \textbf{14973(7984)} & 1.6                 &\textbf{8.8 }                                                                                                \\
HYGRO I* & 40.3(82.3)          & 1035(3313)          & 11.8                &  & \textbf{8.5(12.0)} & \textbf{51.6(70.8)} & \textbf{16.9}        &  & \textbf{237(109) } & \textbf{14126(8054)} & \textbf{ 3.1}        &\textbf{10.6 }                                                                                                \\ \hline
Default       & 45.6(85.8)          & 1234(4078)          & N/A                   &  & 10.2(14.3)         & 64.3(87.8)          & N/A                   &  & 245(96)            & 16247(8624)          & N/A                   & N/A                                                                                                \\
No Cuts       & 123.7(138.3)        & 3228(6788)          & -171.1               &  & 108.5(132.1)       & 337.6(427.3)        & -961.3              &  & \textbf{238(108)}  & 15701(8763)          & \textbf{3.0}        &-376.5                                                                                                 \\
Always   & \textbf{32.8(65.9)} & \textbf{889(1259) } & \textbf{28.1}        &  & 25.9(26.1)         & 88.7(84.4)          & -153.4               &  & 253(91)            & 16704(8766)          & -3.2                 & -42.8
                                                                                                \\
FCN        & 41.8(82.7)          & 1141(2108)          & 8.4                  &  & 41.4(69.3)         & 110.3(120.9)        & -305.1              &  & 252(92)            & 17313(9730)          & -3.0                 &-99.9                                                                                                 \\
FCR        & 51.5(100.4)         & 1742(4901)          & -12.9                &  & 23.5(26.2)         & 88.9(94.7)          & -129.5               &  & 253(86)            & 15410(7715)          & -3.4                 & -48.6

                                                                                                \\
Immediate     & 52.7(91.8)          & 1410(4401)          & -15.5                &  & 13.9(20.7)         & 73.6(95.4)          & -36.2                &  & 238(108)           & 16245(9151)          & 3.0                  & -16.2

                                                                                                \\
Random I      & 42.6(83.6)          & 1106(3553)          & 6.7                  &  & 22.6(23.5)         & 88.1(90.4)          & -121.3               &  & 261(84)            & 16195(7991)          & -6.4                 &-40.3

                                                                                                 \\
Random II     & 41.9(85.4)          & \textbf{995(2280)}  & 8.3                  &  & 19.2(24.1)         & 84.0(88.6)          & -88.0                &  & 245(103)           & 15235(8780)          & -0.0                 &-26.6

                                                                                                 \\ \toprule[6pt]
\end{tabular}
}
\caption{We presented the data for Experiments 1 and 2, which involve evaluations on classical datasets and harder datasets. The top two performances are highlighted in bold. The performance metrics are presented as average (standard deviation) for each test. Solving time improvement (Imprv.) over the default SCIP's stopping strategy and average solving time improvement (Imprv. Avg.) are listed. V and C denote variable and constraint counts, respectively. * represents our approach.}
\end{table*}
\subsection{Experiment Details}

\subsubsection{Dataset\label{Datasets}}
We employed six NP-hard problem datasets as benchmarks, which are categorized into two groups.

Classical NP-hard combinatorial optimization problems widely served as benchmarks, including Multiple Knapsack Problem \cite{knapsack}, Maximum Independent Set (MIS) \cite{MIS}, and Set Covering \cite{Setcover}. These datasets are artificially generated using the methods provided by \cite{ecole}. 

Harder datasets, including Corlat \cite{corlat}, MIK \cite{mik}, and Anonymous \cite{anonymous} dataset. These datasets are related to real-world optimization problems, often employed as benchmarks in MILP research \cite{use_datasets_corlat_et}.
\subsubsection{Setup}
We integrate HYGRO into SCIP, a widely utilized backend for MILP research \cite{Gasse2019ExactCO,paulus2022learning,wang2023learning}, and one of the state-of-the-art open-source solvers.  During solving, we set a 300-second time limit for each MILP instance. HYGRO's training utilizes the ADAM optimizer \cite{Adam} through PyTorch \cite{Torch}. To ensure fairness and replicability, we maintain default SCIP parameter settings, except for those pertaining to the cutting plane stopping strategy. We divided each dataset into training and testing sets with \(75\%\) and \(25\%\) of instances, respectively, and selected the best model trained on the training set for testing.

In the formulation presented in Section \ref{Formulation}, we consider making decisions in each node. In fact, many studies concentrate primarily on decisions made at the root node \cite{Tang2020}. On one hand, this approach will reduce computational overhead; on the other hand, cutting planes added at the root node tend to have a greater impact on solving efficiency. However,  for experimental completeness, in the method HYGRO I, we make decisions for cutting plane stopping at multiple nodes (up to a depth of 2) – whereas all other methods only focus on the root node.

\subsubsection{Performance metrics}

In this study, we assess each method's performance using two key metrics: solving time and PDI. We employ the metric of average improvement to assess the overall performance of each method. Specifically, we calculated the average percentage performance improvement of a method across all datasets in this experiment.

\subsubsection{Baseline}
In our experiments, we employed several baseline strategies, including:
\textbf{(1) Default}: The default cutting plane stopping strategy offered by the SCIP solver, utilizing the Stagnation Round Detection strategy.
\textbf{(2) No Cuts:} This strategy disables the G\&A of cuts.
\textbf{(3) Always:} This strategy performs G\&A of cuts without any conditions until a specific event of SCIP is triggered.
\textbf{(4) FCN:} As discussed in Section \ref{FixedNumber}, this strategy ensures that the number of cuts added in each run does not exceed 200.
\textbf{(5) FCR:} As explained in Section \ref{FixedRound}, this strategy ensures that the number of cutting plane rounds in each run does not exceed 100.
\textbf{(6) Immediate:} This strategy involves setting the value of \(\epsilon\) in Inequality (\ref{stagnation_inequality}) to a very small value of \(10^{-5}\), and as soon as it becomes smaller than \(\epsilon\), cutting plane G\&A is immediately stopped.
\textbf{(7) Random \uppercase\expandafter{\romannumeral1}:} This strategy incorporates a \(0.5\%\) probability of instantly stopping the G\&A of cutting planes before each iteration.
\textbf{(8) Random \uppercase\expandafter{\romannumeral2}:} This strategy involves generating a random integer \( t \) within the range of [0, 30) to serve as the threshold for consecutive stagnation rounds.

Baseline strategies (1)-(5) are provided by SCIP, while the strategies (6)-(8) are designed by us.

\begin{table}[t]

\resizebox{\linewidth}{!}{
\fontsize{24}{28}\selectfont
\begin{tabular}{llcclc}
\toprule[3pt]
                         &  & \multicolumn{2}{c}{MIK}                                                                                                      &  & \multicolumn{1}{l}{\multirow{2}{*}{Imprv.(\%)}} \\ \cline{1-1} \cline{3-4}
Metrics                  &  & Direct                                                        & Indirect (ours)                                                     &  & \multicolumn{1}{l}{}                                 \\ \hline
Solving Time(s)$\downarrow$ &  & 21.5(15.3)                                                  & \textbf{10.9(5.4)}                                                  &  & 49.32

\\
Extra Time(s)$\downarrow$   &  & 7.0(5.8)                                                    & \textbf{1.9(2.6)}                                                   &  & 72.70                                             \\
% Nodes$\downarrow$        &  & \begin{tabular}[c]{@{}c@{}}14059.07\\ (29859.90)\end{tabular} & \begin{tabular}[c]{@{}c@{}}7516.30\\ (12375.48)\end{tabular} &  & 46.53                                             \\
PDI$\downarrow$          &  & 403.0(458.4)                                                & \textbf{239.2(262.7)}                                               &  & 40.62                                             \\
Call Times$\downarrow$   &  & 116.1(108.7)                                                & \textbf{1.0(0.0)}                                                   &  & 99.14                                              \\ \toprule[3pt]
                         &  & \multicolumn{2}{c}{MIS}                                                                                                      &  & \multicolumn{1}{l}{\multirow{2}{*}{Imprv.(\%)}} \\ \cline{1-1} \cline{3-4}
Metrics                  &  & Direct                                                       & Indirect (ours)                                                      &  & \multicolumn{1}{l}{}                                 \\ \hline
Solving Time(s)$\downarrow$ &  & 9.2(4.6)                                                    & \textbf{7.1(4.7)}                                                   &  & 22.18                                             \\
Extra Time(s)$\downarrow$   &  & 2.9(2.4)                                                    & \textbf{1.1(2.2)}                                                   &  & 63.44                                             \\
% Nodes$\downarrow$        &  & 110.03(327.02)                                                & 215.38(879.88)                                               &  & -95.74                                             \\
PDI$\downarrow$          &  & 55.4(36.2)                                                  & \textbf{45.0(34.0)}                                                 &  & 18.82                                             \\
Call Times$\downarrow$   &  & 24.4(2.7)                                                   & \textbf{1.0(0.0)}                                                   &  & 95.90                                         \\ \toprule[3pt]
\end{tabular}
}
\caption{The experimental results of the Ablation Study, the best-performing results are highlighted in bold.}\label{ablation_res}
\end{table}

\subsection{Results}
\subsubsection{Experiment 1: Classical Datasets Performance Experiment}
%%v8.14
To evaluate HYGRO's ability to find efficient stopping conditions for classical MILP problems, we conducted performance comparison experiments with eight baseline strategies. Our HYGRO strategies, including HYGRO I, consistently demonstrate remarkably superior performance, outperforming other strategies in the majority of tests. Although there were two occurrences where baseline strategies slightly outperformed HYGRO, our methods exhibited substantial performance improvements over these two strategies on other datasets. It is worth noting that the default strategy of SCIP, based on the SRD strategy, also proves to be highly effective as a stopping strategy, showing distinct advantages over other baseline strategies. Nevertheless, in comparison to this default strategy, HYGRO achieved an average improvement of approximately \(22\%\), with the highest observed improvement reaching up to \(31\%\) on specific datasets.

\subsubsection{Experiment 2: Harder Datasets Performance Experiment}
To assess HYGRO's performance on more challenging MILP problems related to real-world applications, we conducted comparative experiments with baseline strategies. Our HYGRO strategies consistently demonstrate the highest overall performance. Relative to the SCIP default strategy, which represents one of the best-performing baseline strategies, our methods achieved performance improvement of approximately \(8\%\) and \(10\%\), along with a notable peak improvement of \(17\%\) on specific datasets. HYGRO exhibited a greater capacity for efficiently solving real-world MILP problems on more challenging datasets.

\subsubsection{Experiment 3: Ablation Study\label{ablation}} 

In the previous Section \ref{transformation}, we transformed the problem of learning stopping strategies. We refer to the methods before and after this transformation as \textbf{``Direct"} and\textbf{ ``Indirect"}, respectively. To validate the effectiveness of this transformation, we compared the performance of these two methods when using HYGRO. 

Specifically, in the Direct method, we determine whether to continue the G\&A of cutting planes based on the output of HYGRO. If the \(ratio\) surpasses 0.5, the iteration stops immediately; otherwise, it continues. We evaluated the performance of these two methods on the MIK and MIS datasets. It is crucial to compare the extra computational costs of the two methods, thus we introduce the metric ``Extra Time" to measure the total runtime of HYGRO during the solving process, and ``Call times" to indicate the frequency of invoking HYGRO. It's worth noting that the hardware utilized in this experiment differs from that of the other experiments, resulting in slight performance deviations. Nevertheless, the experimental setup maintains its consistency and fairness.

The results are presented in Table \ref{ablation_res}, from which we can find that the Indirect method exhibits notable advantages in terms of Extra Time metrics, significantly reducing the increased computational overhead from using HYGRO. It still maintains a substantial advantage in overall runtime, indicating that through the transformation, we have learned a more efficient stopping strategy. Even considering only the pure MILP solving time (excluding the Extra Time), the Indirect method still holds a significant advantage. These facts strongly validate the effectiveness of our transformation.
\begin{table}[t]
\resizebox{\linewidth}{!}{
\fontsize{50}{60}\selectfont
\begin{tabular}{lccccccc}
\toprule[4pt]
                           & \multicolumn{2}{c}{HYGRO (ours)}                   & \multicolumn{1}{l}{} & \multicolumn{2}{c}{Default}                 & \multicolumn{1}{l}{} & \multirow{2}{*}{\begin{tabular}[c]{@{}c@{}}Imprv.\\ \%(Time)\end{tabular}} \\ \cline{1-3} \cline{5-6}
Datasets      & Time(s)\scalebox{2}{$\downarrow$}     & PDI\scalebox{2}{$\downarrow$}      &                      & Time(s)\scalebox{2}{$\downarrow$}     & PDI\scalebox{2}{$\downarrow$}      &                      &                                                                               \\ \hline
\begin{tabular}[l]{@{}l@{}}Set Covering I\\(1000V, 1000C)\end{tabular}   &\textbf{\begin{tabular}[c]{@{}c@{}}11.28\\(29.99)\end{tabular}}           & \textbf{\begin{tabular}[c]{@{}c@{}}102.83\\(180.20)\end{tabular}} &&\begin{tabular}[c]{@{}c@{}}14.42\\(24.69)\end{tabular}               &\begin{tabular}[c]{@{}c@{}}133.28\\(155.80)\end{tabular}              &&21.77                                   \\
\begin{tabular}[l]{@{}l@{}}Set Covering II\\(1500V, 1000C)\end{tabular}  &\textbf{\begin{tabular}[c]{@{}c@{}}26.74\\(45.93)\end{tabular}}              &\textbf{\begin{tabular}[c]{@{}c@{}}239.19\\(348.32)\end{tabular}}            &  &\begin{tabular}[c]{@{}c@{}}29.61\\(47.69)\end{tabular}               &\begin{tabular}[c]{@{}c@{}}257.52\\(337.41)\end{tabular}  &&9.69           \\
\begin{tabular}[l]{@{}l@{}}Set Covering III\\(2000V, 1000C)\end{tabular}&\textbf{\begin{tabular}[c]{@{}c@{}}59.76\\(84.28)\end{tabular}}               &\textbf{\begin{tabular}[c]{@{}c@{}}504.76\\(669.90)\end{tabular}}     &  &\begin{tabular}[c]{@{}c@{}}64.23\\(84.52)\end{tabular}               &\begin{tabular}[c]{@{}c@{}}550.48\\(671.31)\end{tabular}     &&6.96            \\ \toprule[4pt]
\begin{tabular}[l]{@{}l@{}}MIS I\\(750V, 3200C)\end{tabular}              &\begin{tabular}[c]{@{}c@{}}28.05\\(24.95)\end{tabular}             &\begin{tabular}[c]{@{}c@{}}105.93\\(84.28)\end{tabular}            &  &\textbf{\begin{tabular}[c]{@{}c@{}}27.08\\(24.70)\end{tabular}} \textbf{}               &\textbf{\begin{tabular}[c]{@{}c@{}}102.80\\(82.50)\end{tabular}}   & &-3.58           \\
\begin{tabular}[l]{@{}l@{}}MIS II\\(1000V, 4200C)\end{tabular}              &\textbf{\begin{tabular}[c]{@{}c@{}}141.13\\(128.76)\end{tabular}}              &\textbf{\begin{tabular}[c]{@{}c@{}}394.57\\(393.85)\end{tabular}}            &  &\begin{tabular}[c]{@{}c@{}}147.68\\(126.69)\end{tabular}              &\begin{tabular}[c]{@{}c@{}}409.31\\(391.22)\end{tabular} &&4.43           \\
\begin{tabular}[l]{@{}l@{}}MIS III\\(1500V, 6400C)\end{tabular}             &\textbf{\begin{tabular}[c]{@{}c@{}}724.34\\(172.01)\end{tabular}}            & \textbf{\begin{tabular}[c]{@{}c@{}}1965.79\\(807.40) \end{tabular}}&     &\begin{tabular}[c]{@{}c@{}}728.61\\(162.21)  \end{tabular}  &\begin{tabular}[c]{@{}c@{}}1972.64\\(802.92)  \end{tabular}             &&0.58              \\ \toprule[4pt]
\end{tabular}
}
\caption{The generalization ability of HYGRO, the best-performing results are highlighted in bold. In the MIS dataset, the count of constraints (C) is an approximate value.}\label{Generalization_Experiment_Results}
\end{table}
\subsubsection{Experiment 4: Generalization Ability}
To evaluate the ability of HYGRO to generalize across various sizes of MILPs, we conducted generalization experiments. We trained HYGRO on the dataset used in Experiment 1 and then assessed the performance of HYGRO and SCIP's default strategy on more challenging datasets within the same domain. Similar settings are commonly employed to evaluate the generalization ability of data-driven models for MILP, as seen in studies like \cite{Gasse2019ExactCO,wang2023learning}. Each instance has an 800-second time limit.

The results of these experiments are elaborated in Table \ref{Generalization_Experiment_Results}.  It is evident that, apart from one occurrence where the default strategy of SCIP slightly outperformed our approach, HYGRO consistently performed better, achieving up to a 21\% improvement in performance. As evident from the results, our approach significantly surpasses the heuristic default stopping strategy when applied to larger datasets. This highlights the robust generalization capacity of HYGRO.

\begin{figure}[t]
  \centering
  \includegraphics[width=0.24\linewidth]{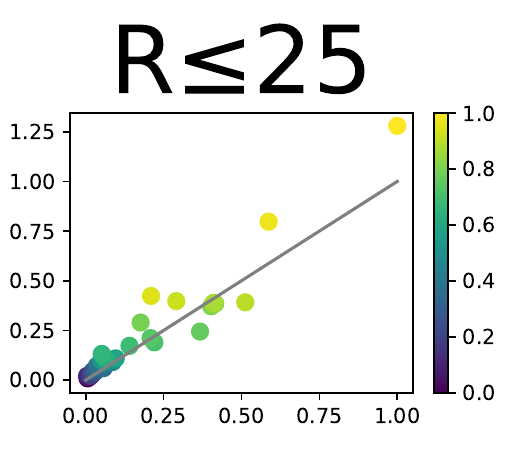}
  \includegraphics[width=0.24\linewidth]{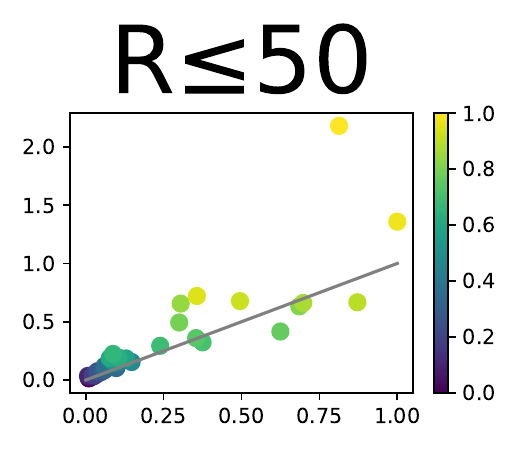}
  \includegraphics[width=0.24\linewidth]{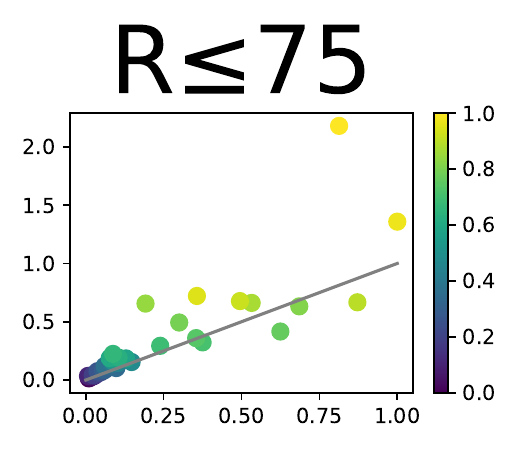}
  \includegraphics[width=0.24\linewidth]{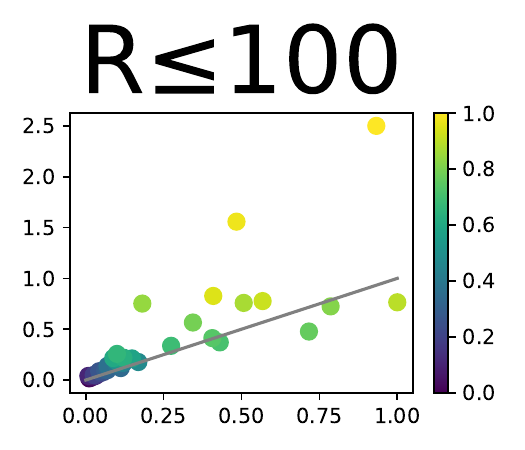}  

  \caption{ The  y-axis shows HYGRO's solving time, and the x-axis indicates the best performance within the specified range of rounds (R). Points below the gray dashed line signify superior performance for HYGRO. }
  \label{fig:performance}

\end{figure}

% \begin{figure}[t]
%   \centering
%   \begin{subfigure}[b]{0.48\linewidth}
%     \centering
%     \includegraphics[width=\linewidth]{MIK_VIS25.pdf}
%   \end{subfigure}
%   \begin{subfigure}[b]{0.48\linewidth}
%     \centering
%     \includegraphics[width=\linewidth]{MIK_VIS50.pdf}
%   \end{subfigure}
  
%   \begin{subfigure}[b]{0.48\linewidth}
%     \centering
%     \includegraphics[width=\linewidth]{MIK_VIS75.pdf}
%   \end{subfigure}
%   \begin{subfigure}[b]{0.48\linewidth}
%     \centering
%     \includegraphics[width=\linewidth]{MIK_VIS100.pdf}
%   \end{subfigure}
  
%   \caption{The data points below the gray dashed line indicate instances where HYGRO's performance surpasses optimal performance from exhaustive search within a specific range.}
%   \label{fig:performance}
% \end{figure}

\subsubsection{Experiment 5: Visualization for Strategy Performance}  To present an intuitive portrayal of the performance of HYGRO, we conducted visualization experiments. For instances in the MIK test dataset, we conducted 100 solving runs. These run data were divided into four distinct sets: $S_1, S_2, S_3,$ and $S_4$. Each set employs a maximum number of cutting plane rounds set at 25, 50, 75, and 100, correspondingly. For example, $S_2$ encompasses the solving performance data for all instances with cutting plane rounds varying from 1 to 50. MILPs in the MIK test dataset are represented by four data points $\{(P_{H}, P_{S_i}^*),i\in\{1,2,3,4\}\}$, plotted on four individual scatter plots, respectively. Here, $P_{H}$ denotes the solving time of HYGRO, while $P_{S_i}^*$ represents the best solving time within $S_i$.

% We conducted visualization experiments to demonstrate the performance. For each instance in the MIK test dataset, we performed 100 solving iterations. In each run, we set the maximum number of cutting plane rounds to $i$. Based on this maximum number of rounds, we divided the 100 runs into four sets: $S_1, S_2, S_3,$ and $S_4$, corresponding to max=25, 50, 75, and 100, respectively. For instance, $S_2$ contains the solving performance data for all MILP instances in test datasets with rounds ranging from 1 to 50. Afterwards, for each MILP instance, it corresponds to four data points $(P_{HYGRO}, P_{S_i}^*)$, where $i \in [1,2,3,4]$, $P_{HYGRO}$ represents the solving performance using our model, and $P_{S_i}^*$ represents the best solving performance in the $S_i$. We plot these data points on a scatter plot. 

The results are visualized in Figure \ref{fig:performance}. It is evident that in most cases, the scatter plots cluster around the vicinity of the gray dashed line, indicating that even when expanding the search range to 100, the performance achieved by HYGRO remains comparable to the optimal values obtained through exhaustive search. Moreover, in certain instances, HYGRO even demonstrates notable advantages. It is important to note that conducting such exhaustive searches comes with a substantial computational cost. This clearly illustrates that HYGRO has effectively learned an efficient stopping condition. 

\section{Conclusion}
In this paper, we focus on the stopping strategy for the cutting plane method. Motivational experiments demonstrate the significance of stopping strategies and highlight the crucial role of learning-based methods. To learn more intelligent stopping strategies, we introduce an innovative transformation to learn a simple and efficient stopping condition and propose HYGRO to learn such conditions. Extensive experiments demonstrate the remarkable performance of HYGRO and provide evidence for the effectiveness of our transformation. We believe HYGRO offers a new perspective on improving the solving efficiency of MILP.

\section*{Acknowledgements}
The authors would like to thank the associate editor and all the anonymous reviewers for their insightful comments. This work was supported in part by National Key R\&D Program of China under contract 2022ZD0119801, National Nature Science Foundations of China grants U19B2026, U19B2044, 61836011, 62021001, and 61836006.

% \bibliography{aaai24}

\end{document}